\documentclass[10pt,twocolumn,letterpaper]{article}

\usepackage{cvpr}
\usepackage{times}
\usepackage{epsfig}
\usepackage{graphicx}
\usepackage{amsmath}
\usepackage{amssymb}

\usepackage{times}
\usepackage{subfigure} 
\usepackage{helvet}
\usepackage{courier}
\usepackage{amsthm}
\usepackage{amsopn}
\usepackage{amsmath,bm}
\usepackage{array}
\usepackage{booktabs}
\usepackage{algorithm}
\usepackage{algorithmic}
\usepackage{enumerate}
\usepackage{color}
\usepackage{multirow}

\newcommand{\sgn}{\text{Sign}}

\usepackage[pagebackref=true,breaklinks=true,letterpaper=true,colorlinks,bookmarks=false]{hyperref}

\cvprfinalcopy 

\begin{document}

\title{Widening and Squeezing: Towards Accurate and Efficient QNNs}

\author{Chuanjian Liu, Kai Han, Yunhe Wang, Hanting Chen, Qi Tian, Chunjing Xu\\
Huawei Noah's Ark Lab\\
{\tt\small \{liuchuanjian, kai.han, yunhe.wang, chenhanting, tian.qi1, xuchunjing\}@huawei.com}
}

\maketitle

\begin{abstract}
   Quantization neural networks (QNNs) are very attractive to the industry because their extremely cheap calculation and storage overhead, but their performance is still worse than that of networks with full-precision parameters. Most of existing methods aim to enhance performance of QNNs especially binary neural networks by exploiting more effective training techniques. However, we find the representation capability of quantization features is far weaker than full-precision features by experiments. We address this problem by projecting features in original full-precision networks to high-dimensional quantization features. Simultaneously, redundant quantization features will be eliminated in order to avoid unrestricted growth of dimensions for some datasets. Then, a compact quantization neural network but with sufficient representation ability will be established. Experimental results on benchmark datasets demonstrate that the proposed method is able to establish QNNs with much less parameters and calculations but almost the same performance as that of full-precision baseline models, \eg $29.9\%$ top-1 error of binary ResNet-18 on the ImageNet ILSVRC 2012 dataset.
\end{abstract}

\section{Introduction}

Deep neural networks especially the convolutional neural networks get state of the art
performance in various computer vision applications, such as image classification~\cite{AlexNet, VGGnet, GoogleNet, ResNet, LeNet} and object detection~\cite{RCNN, fasterRCNN, SSD}, semantic segmentation~\cite{FCN, deeplab_v1, maskrcnn}, etc. Applications embedded on mobile devices can benefit from the advantages of low latency, better privacy and offline operation. However, deploying deep models on resource constrained mobile devices is challenging due to high memory and computation cost.  Motivated by this demand, researchers have proposed many model compression and accelerate methods to improve the applicability of the learned deep models, e.g., pruning~\cite{prun1, prun2, pruning, pruning15,liu2019learning}, quantization~\cite{quan1,quan2,shen2019searching}, decomposition~\cite{matrix1, matrix3} and lightweight structure~\cite{mobilenet_v1, shufflenet_v1, squeezenet}.

One of the most widely used model compression approaches is quantization. To reduce the complexities of deep CNNs, a number of recent works have been proposed for quantizing weights or activations or both of them. Wherein, binary neural networks with weights and activations constrained to $+1$ or $-1$ have many advantages. In contrast to non-binary networks, binary network needs smaller disk memory and replace most arithmetic operations with bit-wise operations XNOR and POPCOUNT, which is power-efficiency and drastically reduce memory size and accesses at run-time. 

BinaryNet~\cite{binary} introduced BNNs with binary weights and activations at run-time and how to compute the parameters gradients at train-time. Its accuracy drops significantly in contrast to full precision nets on ImageNet~\cite{ImageNet}. XNOR-Net~\cite{XNORNet} used the real-valued version of the weights and activations as a key reference for the binarization process. The classification accuracy of a Binary-Weight-Network version of AlexNet is the same as the full-precision AlexNet. But the performance of XNOR-Net drops a lot. BinaryConnect~\cite{binaryconnect} trained a DNN with binary weights during the forward and backward propagations. It achieved state-of-the-art results on small datasets, but experiments~\cite{XNORNet} show that this method is not very successful on large-scale datasets. BNN$+$~\cite{bnn+} proposed a new regularization function that encourages training weights around binary values to reduce this accuracy gap. \cite{binaryensemble} leverages ensemble methods to improve the performance of BNNs with limited efficiency cost. 

Although the aforementioned methods have made tremendous efforts to increase the performance of binary neural networks by adjusting their architectures (\eg change the order of ReLU and batch normalization~\cite{LQNet}) and adding more regularizations~\cite{bnn+}. In fact, most algorithms for learning binary neural networks can be regarded as a binary feature embedding task in which dimensions of binary features and original high-bit features are exactly the same. The representation ability of features in the binary space will be definitely lower than that of features in the high-bit space, if there are no obvious redundancy in original features. We are therefore motivated to increase the number of binary filters to a suitable value for obtaining binary features with the same effectiveness.

To this end, we first provide two experiments to show the representation ability of binary features. Firstly, to discover the intrinsic representations of deep features via binary feature embedding, features in the given full-precision neural networks will be first projected into a high-dimensional binary space using an orthogonal transformation for retaining their pair-wise Euclidean distances. Then, redundancy in the original features are recognized through a learned selection mask. Based on the obtained compact binary features, we re-configure the neural network with an acceptable increment on its number of filters. Experiments on benchmarks demonstrate that, binary neural networks established using the proposed method are able to achieve the similar performance as that of full-precision baseline models with significantly lower memory usage and OPs. Secondly, we show the accuracy of widened binary networks is higher in contrast to deepened binary networks. Then we proposed the quantization methods and new network design architecture.

To summarize, our main contributions are as follows:
\begin{itemize}
	\item We analyze the feature transformation from full precision to low bit representations, and the optimization results prove that the transformation process is effective. Then we use another experiment to show that widened network gets more benfit in contrast to deepened network.
	\item The quantization method are proposed. Besides, network pruning are use to search for the efficient and accurate quantized network architecture. Knowledge distillation is used to improve the performance of quantized network.
	\item Experiments on benchmark classification and detection dataset verify the effectiveness of proposed method.
\end{itemize}

In the rest of the paper, we first revisit the related works in this area and describe the preliminaries and motivation. Next we introduce our approach to learn accurate and efficient quantization networks. Then the experiments are presented and analyzed to show the effectiveness of our method. Finally, the conclusion is made. 

\section{Related Works}

This paper solves the limit representation ability of quantization networks by use more features. Previous studies have focused on designing new network architecture or innovative quantization function or finding the best distribution of quantized values. In this section, we revisit the existing methods for establishing compact models, including network quantization, network pruning and knowledge distillation.

\subsection{Network Quantization}
BinaryConnect~\cite{binaryconnect} directly optimizes the loss of the network with weights $W$ replaced by $sign(W)$, and it approximates the sign function with the "hard tanh" function in the backward process to
avoid the zero-gradient problem. The binary weight network (BWN)~\cite{binary} adds scale factors for the weights during binarization. Ternary weight network (TWN)~\cite{li2016ternary} introduces ternary weights and achieves improved performance. XNOR-Net~\cite{XNORNet} proposed to add a real-valued scaling factor to each output channel of a binary convolution. Trained ternary quantization (TTQ)~\cite{TTQ} proposes learning both ternary values and scaled gradients for 32-bit weights. DoReFa-Net~\cite{dorefanet} proposes quantizing 32-bit weights, activations and gradients using different bit widths. Gradients are approximated by a customized form based on the mean of the absolute values of the full-precision weights. In~\cite{pruning}, pruning, quantization and Huffman coding are used to compress model. Bi-Real net~\cite{bireal} connects the real activations (after the 1-bit convolution and/or BatchNorm layer, before the sign function) to activations of the consecutive block, through an identity shortcut to enhance representational capability. ABC-Net~\cite{ABCnet} and binary ensemble method~\cite{binaryensemble} use more more convolutions operations per layer to improve accuracy. Although these works have made great progress, the performances of low-bit quantization networks, especially 1-bit neural networks are still much worse than the full-precision counterparts.

\subsection{Network Pruning}
Network pruning is an effective technique to compress and accelerate CNNs, and thus allows us to deploy efficient networks on hardware devices with limited storage and computation resources. Structured pruning methods~\cite{softprune, li2016pruning, channelprune, pruning15} target the pruning of convolutional filters or whole layers, and thus the pruned networks can be easily developed and applied. For example, Liu et al.~\cite{netslimming} leveraged a $\ell_1$ regularization on the scale factors to select channels. He et al.~\cite{filterprune} utilized a geometric median-based criterion to cut out unimportant filters. In this paper, we utilize the network pruning technique to slim the widened QNNs to achieve lower memory and computational cost.

\subsection{Knowledge Distillation}
Knowledge distillation is one of the most popular solutions for model compression. The idea is to improve the performance of small model with transferred soft targets provided by the large model. Hinton et al.~\cite{Distill} proposed the knowledge distillation approach to compress the knowledge of a large and computational expensive model to a single computational efficient neural network. Since then, knowledge distillation has been widely adopted and many methods are proposed. For example, Romero et al.~\cite{FitNet} proposed FitNet, which extracted the feature maps of the intermediate layer as well as the final output to teach the student network. After that, Zagoruyko et al.~\cite{attentiontransfer} defined Attention Transfer based on attention maps to improve the performance of the student network.

In this paper, we join the pruning and distillation with quantization networks. The network pruning methods are used to find the efficient quantized network. Then we use knowledge distillation to improve the accuracy of compact quantization networks.

\section{Preliminaries}

Firstly, binary embedding of high-dimensional data requires long codes to preserve the discriminative power of the input space. So we try to find the lower bounds of binary embedding for the feature maps in one layer of CNN. Secondly, we show the performance of widened binary network is better than deepened binary networks.

\subsection{Binary Feature Embedding}
Our goal is to find the minimum dimensionality (\ie the number of filters) of binary neural networks for preserving the performance of full-precision neural networks.

For an arbitrary convolutional layer in a pre-trained deep neural network, the convolution operation for a given instance can be formulated as
\begin{equation}
X^\top F+b = Y,
\label{Fcn:conv}
\end{equation}
where $X\in\mathbb{R}^{wh\times ck^2}$ is the input data after converting to a matrix according to filter size and parameters in this layer (\ie the original images or activations of the previous layer), $F\in\mathbb{R}^{ck^2\times n}$ stacks $n$ convolution filters, and $Y\in\mathbb{R}^{wh\times n}$ is the output feature maps, $w$ and $h$ are the width and height of out feature maps, respectively. $n$ is the output channel number, and $b$ is the bias term, which is often eliminated for simplicity.

For the neural network binarization problem, we denote the approximated binary feature maps as $\tilde{Y} = \phi(Y)\in\mathbb{R}^{wh\times m}$, where $m$ is the number of filters in the binarize layer, and $\phi(\cdot)$ could be either a linear~\cite{LPP} or non-linear transformation~\cite{LLE, isomap}. Commonly, we can utilize a linear transformation $P$ to accomplish this as suggested in~\cite{CBE,LPP}, \ie $\tilde{Y} = YP^\top$, where $P\in\mathbb{R}^{m\times n}$. Thus, the binarization on feature maps $Y$ can be formulated as
\begin{equation}
\min_{P,B} \frac{1}{2}|| YP^\top-B||_F^2,
\label{Fcn:binary_}
\end{equation}
where $B\in\{-1,+1\}^{wh\times m}$ is the binarized feature maps, $||\cdot||_F$ is the Frobenius-norm for matrices. Note that, the number of filters $m$ in the binary network could be either larger or smaller than $n$, which will be discussed in the follow. 

The above function only force features (or activations) in the given convolutional layer are binary, which does not inherit the functionality of features learned on massive training data. Therefore, we propose to retain the relationship between features of every two samples, which is commonly the most important characteristic in visual recognition tasks such as image classification, detection, and segmentation, \ie
\begin{equation}
\min_{P,B} \frac{1}{2}|| YP^\top-B||_F^2+\frac{\gamma}{2}||\mathcal{D}(Y)-\mathcal{D}(B)||_F^2,
\label{Fcn:objD}
\end{equation}
where $\mathcal{D}(\cdot)$ calculates the Euclidean distances between features of all samples in the training set. Since the number of samples in training set is usually very large (\eg, 1M in the ImageNet~\cite{ImageNet}), $\mathcal{D}(Y)$ is an extremely huge matrix, \eg $10^6\times 10^6$, which cannot be efficiently optimized. Fortunately, if $P$ is an orthogonal matrix when $P$ is square matrix or $P^\top P=I$, the Euclidean distances between any two samples features can be completely preserved. Given two features $Y_i$ and $Y_j$ generated using the original network, we have 
\begin{equation}
\small
\begin{aligned}
||Y_i P^\top-Y_j P^\top||_F^2 &= Tr(Y_i  P^\top P Y_i^\top) - 2Tr(Y_i P^\top P Y_j^\top) \\
&~~~~~~~~+ Tr(Y_j P^\top P Y_j^\top)  \\
&=Tr(Y_i Y_i^\top) - 2Tr(Y_iY_j^\top) + Tr(Y_j Y_j^\top) \\
&=||Y_i -Y_j||_F^2.
\end{aligned}
\label{Fcn:dist}
\end{equation}

Therefore, we reformulate Fcn.~\ref{Fcn:objD} as
\begin{equation}
\min_{P,B} \frac{1}{2}|| YP^\top-B||_F^2+\frac{\gamma}{2}||P^\top P-I||_F^2,
\label{Fcn:objP}
\end{equation}
where $I$ is an $n\times n$ identity matrix, and $\gamma$ is a hyper-parameter for balancing two terms in the above function. Then, we can utilize Fcn.~\ref{Fcn:objP} to binarize the given full-precision network for maintaining its performance.

Further, we want to find the lower bounds of binary embedding for the feature maps, which means the column sparsity of binary representation $B$. So we introduce mask $M\in\{0, 1\}^{m}$ which is used to select features of $B$. $M\circ B$ represent the multiplication of element in $M$ and column in $B$. The $L_1$-norm of mask $M$ can be used to find the lower bounds. Finally, the follow function Fcn.~\ref{Fcn:final} is used to get the optimal binary embedding of features.
\begin{equation}
\min_{P,B,M} \frac{1}{2}|| YP^\top-M\circ B||_F^2+\frac{\gamma}{2}||P^\top P-I||_F^2+\beta||M||_1.
\label{Fcn:final}
\end{equation}

Alternative optimization is use to solve Fcn.~\ref{Fcn:final}: 

\textbf{Solve $B$.} Since elements in binary features $B$ are independent, which could be simply obtained by 
\begin{equation}
B = \sgn(YP^\top),
\end{equation}
where $\sgn(\cdot)$ outputs the signs of the input data.

\textbf{Solve $M$.} For fixed binary variables $B$, and projection $P$, the optimization of $M$ can be formulated as:
\begin{equation}
\mathcal{L}(M) = \frac{1}{2}||YP^\top-M\circ B||_F^2+\alpha ||M||_{1},
\label{Fcn:solveM}
\end{equation}
which aims to eliminate some column with larger reconstruction errors.

\textbf{Solve $P$.} For the solved mask $M$ and binary variables $B$, the loss function for optimizing the projection matrix $P$ can be written as:
\begin{equation}
\mathcal{L}(P) = \frac{1}{2}||YP^\top-M\circ B||_F^2+\frac{\gamma}{2}||P^\top P-I||_F^2,
\label{Fcn:solveP}
\end{equation}

We take use of vgg-small on CIFAR-10 and only optimize the final convolutional features, results are in Table.~\ref{Tab.1}. Mini-batch SGD is used to solve $P$ and $M$. To fully excavate the representation ability, we set $m$ be $8$ times of $n$ initially. 
In Table.~\ref{Tab.1}, the optimized number of channels are shown in the third line. In the low level layers, more features are expected and less feature are needed in the high level layers. Then we retrain the optimized binary nets and get $92.44\%$ accuracy.

\begin{table}
	\begin{center}
		\begin{tabular}{|l|c|c|c|c|c|c|}
			\hline
			Methods & 2 & 3 & 4 & 5 & 6 & acc \\
			\hline\hline
			baseline & 128 & 256 & 256& 512& 512& 0.9394 \\
			\hline
			binary   & 410 & 332 & 614& 420&  25& 0.9244 \\
			\hline
		\end{tabular}
	\end{center}
	\label{Tab.1}
	\caption{Results of binary embedding. The number in the first row means the layer of vgg-small.}
\end{table}

Though this optimization method provides one method to find the number of binary features, it is hard to optimize and we have to optimize all convolutional features layer by layer. So one simple but efficient method is expected.

\subsection{Widen or Deepen?}
We did experiments on different width binary networks and depth binary networks on CIFAR-10. The full precision ResNet-20 is chosen as the baseline. We use $\{1, 2, 3, 4, 8\}$ times width binary ResNet-20, and binary ResNet-32, ResNet-56 and ResNet-110 are chosen as different depth networks. 

\begin{figure}[t]
	\begin{center}
		\includegraphics[width=1.0\linewidth]{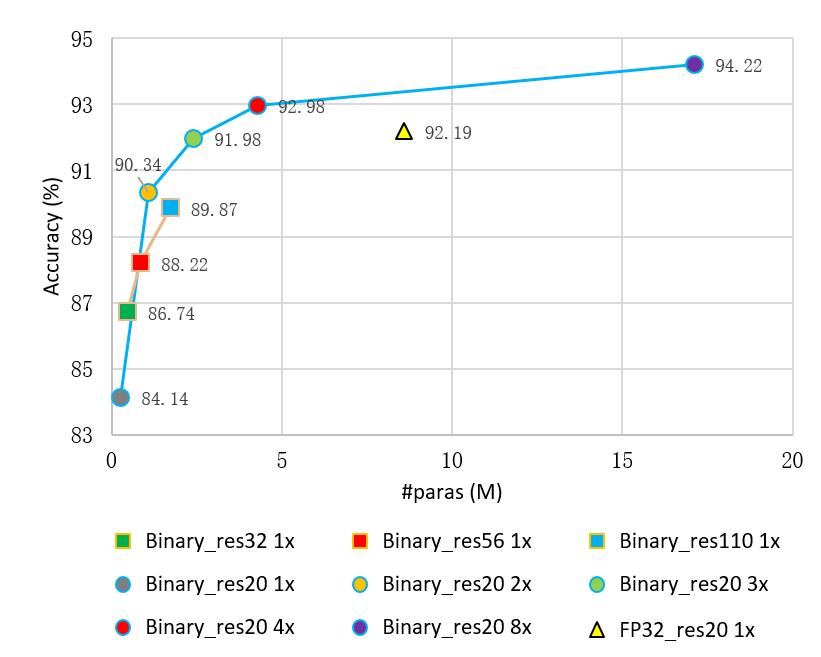}
	\end{center}
	\caption{An illustration of the widened binary networks and deepened binary networks.}
	\label{fig:toy}
\end{figure}

In figure~\ref{fig:toy}, the performance of $2$-times width ResNet-20 with less parameters is better than ResNet-110. $3$-times width binary ResNet-20 uses much less parameters and achieves almost the same accuracy as full-precision ResNet-20. The accuracy of $4$ and $8$-times width model are further higher with more channels in each layer. We conjecture the reason lies on that the representation ability of binary features is lower than the full precision features, so we expected to use more binary features to improve performance.

\section{Approach}

According to above experiments, we expected to use as less as quantized features to get high precision model. So we use the net slimming method on widened quantization networks to search for efficient architectures.

\subsection{Implementation of Quantization Layer}

In this paper, Weight and activation are quantized respectively. Without loss of generality, we assume $W$ be the real weight and $A$ be the real activation, and $Q_W$ be the quantized weight except that binary weights are expressed by $B_W$, quantized activation is represented with $Q_A$. 
For binary weight, the common binary method in XNOR-net are used. The optimal estimation for $W\approx \alpha B_W$ is $\alpha^* = \overline{\lVert W\rvert_1}$ and $B_W$$=$$Sign(W)$. Where $Sign(\cdot)$ returns the sign of input. $\overline{\lVert W\rvert_1}$ is the mean value of weight. The implementation uses Straight-Through Estimator (STE) to back propagate the gradients.

The process of $n$-bits ($n\neq 1$) quantization of weight contains three steps:
\begin{itemize}
	\item[1] $W=\frac{\tanh(W)}{2\cdot\max(|\tanh(W)|)}+0.5$, the value of weight are project to $[0, 1]$;
	\item[2] $W=\frac{round(W\cdot scale)}{scale}$, where $scale=2^n-1$, the quantized weight are in $\{0, \frac{1}{2^n-1}, \frac{2}{2^n-1}, \cdots, 1\}$;
	\item[3] $Q_W = 2\cdot W-1$, the value are project to $[-1, 1]$.
\end{itemize}
In step $2$, we also use the STE for gradient back-propagation.

For $n-$bits quantization of activation, we first clamp the value in $X$ to $[0, 1]$, and then use $\frac{round(X\cdot scale)}{scale}$ to get the quantized value. Where $scale=2^n-1$. The back-propagation also take use of STE.

\begin{figure}[t]
	\begin{center}
		\includegraphics[width=0.9\linewidth]{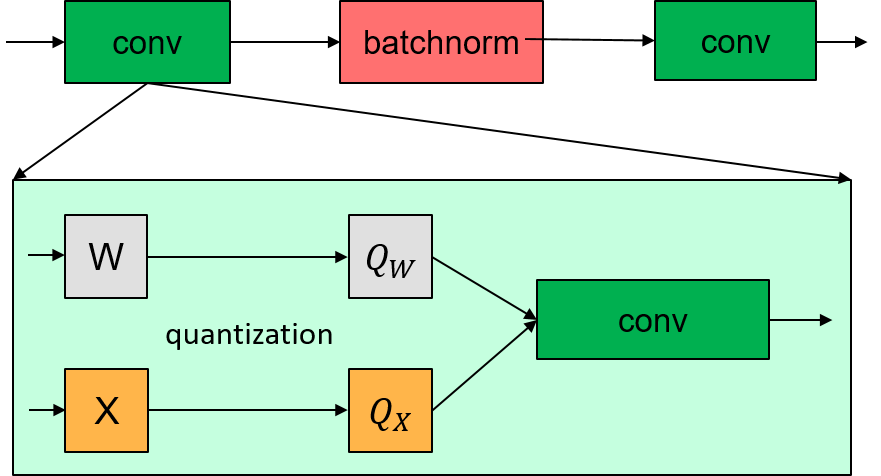}
	\end{center}
	\caption{Typical block of quantized networks.}
	\label{fig:block}
\end{figure}

The typical block are shown in Figure~\ref{fig:block}. Generally, the first layer and the last layer are not quantized, because there are too much information loss if the input images are quantized and metric loss in output. During training, the features after batch normalization layer and the kernel weight are quantized respectively. Then they are transferred to the convolutional layer. After training, the binary weight are saved. When inference, only the feature are need to re-quantization to compute the result.

\subsection{Network Architecture Design}

In the preliminary experiments, we illustrate that the representation ability of binary feature is not powerful and the best solution is to widen the binary network. Then some problem appears, how many times should we widen the network?

Network pruning is widely used for reducing the heavy inference cost of deep models in low-resource settings. The pruned architecture itself has both low FLOPs and high accuracy. Besides, pruning method can be seen as an architecture search paradigm which is used to find the best and efficient architecture.

In order to use the advantage of network pruning, we choose at least $4$ times width to get better accuracy in contrast to the standard network. Then we use the network slimming~\cite{netslimming} method to prune networks and get efficient models. In particular, network slimming imposes a sparsity penalty on the scale factors $\gamma$ in batch normalization layers. The training objective during pruning is
\begin{equation}
\mathcal{L}_{\mathrm{pruning}}=\mathcal{L}_0 + \lambda\|\gamma\|_1,
\end{equation}
where $\mathcal{L}_0$ is the original loss function for specific task, \eg cross entropy loss for classification task and mean square error (MSE) loss for regression task, and $\lambda$ is the sparsity regularization hyper-parameter. During training, insignificant channels are automatically identified and pruned afterwards, yielding thin and compact models with comparable accuracy. Then we retrain or fine-tune the compact model to get high accuracy.

Knowledge distillation is one model compression method in which a small model is trained to mimic a pre-trained, larger model (or ensemble of models). In distillation, knowledge is transferred from the teacher model to the student by minimizing a loss function in which the target is the distribution of class probabilities predicted by the teacher model. Denote the pre-softmax outputs (\ie logits) of teacher model and student model as $o_T$ and $o_S$ respectively, and the softmax predictions as $p_T=\mathrm{softmax}(o_T)$ and $p_S=\mathrm{softmax}(o_S)$. The knowledge distillation loss can be formulated as:
\begin{equation}
\mathcal{L}_{\mathrm{KD}}=\mathcal{H}(y,p_S)+\mu\mathcal{H}(p_T^\tau,p_S^\tau),
\end{equation}
where $\mathcal{H}(\cdot,\cdot)$ is the cross entropy loss, $y$ is the groud-truth one-hot label vector and $\mu$ is the trade-off hyper-parameter to balance these two terms. Moreover, $p_T^\tau$ and $p_S^\tau$ are the softened predictions of teacher model and student model:
\begin{equation}
p_T^\tau=\mathrm{softmax}(\frac{o_T}{\tau}), ~ 
p_S^\tau=\mathrm{softmax}(\frac{o_S}{\tau}),
\end{equation}
where $\tau$ is the temperature parameter. We take advantage of knowledge distillation to improve the accuracy of the compact models. The teacher model could be the full-precision model or widened binary network.

\section{Experiments}

In this section, we will implement experiments to validate the effectiveness of the proposed quantization method on several benchmark classification image datasets and one detection task. Experimental results will be analyzed to further help to understand the benefits of the proposed approach.

\subsection{Datasets and Settings}

To verify the effectiveness of the proposed quantization method, we conduct experiments on several benchmark visual datasets, including CIFAR-10~\cite{cifar10}, CIFAR-100~\cite{cifar10}, ImageNet ILSVRC 2012 dataset~\cite{ImageNet}, and PASCAL VOC0712 object detection benchmark~\cite{pascal}. CIFAR-10 dataset is utilized for analyzing the properties of the proposed method, which consists of $60,000$ colour images in $10$ classes, with $50,000$ training images and $10,000$ test images. CIFAR-100 dataset has the same number of images except that is has $100$ classes. A common data augmentation scheme including random crop and mirroring are adopted. ImageNet is a large-scale image dataset which contains over $1.2M$ training images and $50K$ validation images belonging to $1,000$ classes. The common data preprocessing strategy including random crop and flip are applied during training. We also conduct object detection experiments on PASCAL VOC0712 dataset. Following common practice, we train models on trainval set (about $16,500$ images)  and evaluate on the VOC07 test split with $4,952$ images.

We make experiments on one or several widths from $\{1,2,3,4,5,8\}$ for each $n$-bit quantization networks. When $width$$=$$1$, the quantized network gets the same width with standard network. For all baseline experiments, we set weight decay as $5e$-$5$. For CIFAR-10 and CIFAR-100, ResNet-20~\cite{ResNet} is selected as the baseline network structure. ResNet-18~\cite{ResNet} and VGG16~\cite{VGGnet} is use to test the performance of ImageNet. SSD~\cite{SSD} detection model with VGG16~\cite{VGGnet} as backbone is used to verify the performance on detection task.

\subsection{CIFAR-10}

We do experiments on $1$ bit and $4$ bit quantization networks. In Table~\ref{Tab.2}, the baseline ($n$$=$$32$ bit) top-1 accuracy is $92.19\%$. For $1$ bit binary network, the accuracy become higher with the increase of network width. When $width$$=$$4$, the accuracy of binary network surpasses the baseline. For $4$ bit quantization network, the accuracy surpasses the baseline when $width$$=$$2$. This result shows that more features are needed when quantization networks get less bits.

\begin{table}
	\begin{center}
		\begin{tabular}{|c|c|c|c|}
			\hline
			$bits$ & $width$ & $paras(M)$ & $accuracy$  \\
			\hline\hline
			32   & 1 & 0.27& 92.19 \\
			\hline
			1    & 1 & 0.27& 84.14 \\
			\hline
			1    & 2 & 1.07&90.34 \\
			\hline
			1    & 3 & 2.41&91.98 \\
			\hline
			1    & 4 & 4.28&92.98 \\
			\hline
			1    & 8 & 17.12&94.22 \\
			\hline
			4    & 1 & 0.27&90.23 \\
			\hline
			4    & 2 & 1.07&93.01 \\
			\hline
			4    & 4 & 4.28&94.39 \\
			\hline
		\end{tabular}
	\end{center}
	
	\caption{Results of different quantization bits and widths on CIFAR-10.}
	\label{Tab.2}
	\vspace{-1em}
\end{table}

\begin{figure}[t]
	\begin{center}
		\includegraphics[width=1.0\linewidth]{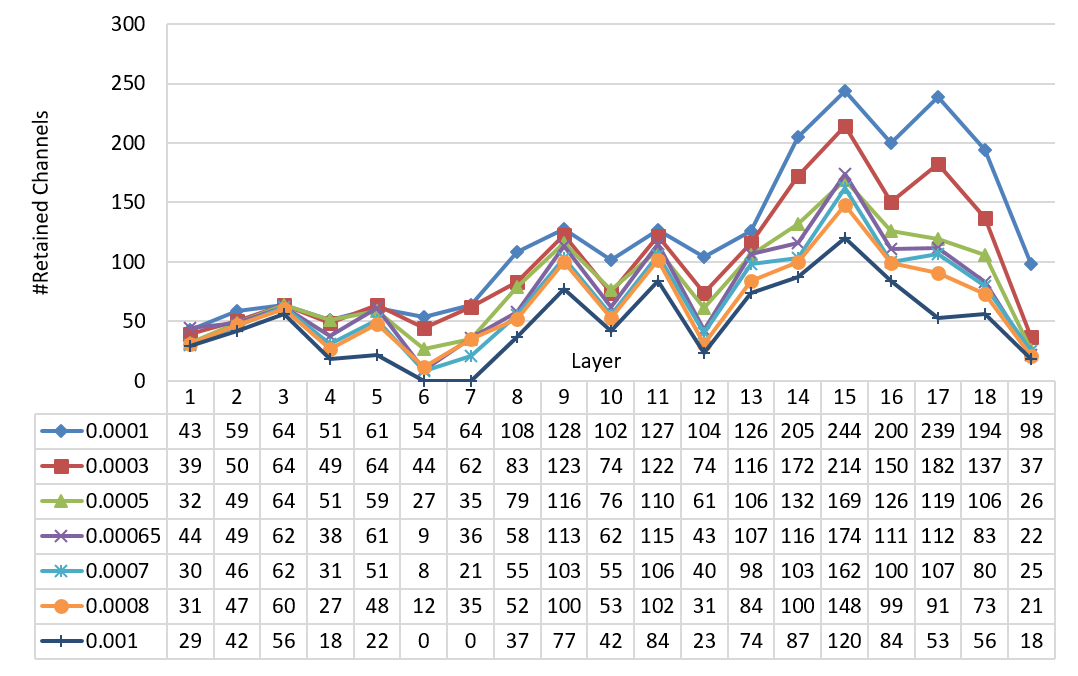}
	\end{center}
	\caption{Network slimming results with different batch norm scale regularization $\lambda$ on CIFAR-10.}
	\label{fig:slimming}
\end{figure}

Then we want to use as less quantized features as possible to get high accuracy. Widely used network pruning method is one effective approach to get light-weight but with high accuracy network. Beside, networks pruning can be seen as one network architecture search method which is used to find efficient network. In this paper, we take use of network slimming method~\cite{netslimming} to get small but accurate quantized networks. In Figure~\ref{fig:slimming}, the slimming results of different batch norm scale regularization factors are illustrated for ResNet-20. In this experiments, we use binary network and set all $width$$=$$4$, and the threshold of slimming is set to $0.01$. From Table~\ref{Tab.3} and Figure~\ref{fig:slimming}, we can find that with the increase of batch norm scale regularization factor, there are more channels been pruned and the accuracy become lower. For each residual block, there are two convolutional layers. The first convolutional layer has more channels been pruned, the reason lies that the shortcut connection prevents too many information been discarded. More shortcut connections can improve the performance of quantization networks.

ResNet has shortcut connection which is identity mapping. For each residual block, the input layer has the same number of channels with the output layer. However, the pruned results usually don't have this characteristic. In our implementation of pruned residual block, we use the channels of output layer as the reference value and try to pad or shrink the channels of input layer to make sure they have the same number of channels. Due to the difference between pruned channels of input layer and output layer, the reconstructed pruned quantization network can no longer use the pre-trained weights. We retrain all pruned quantization resnet to get $acc_R$. $acc_P$ is the result when all pruned channels are set to $0$.

\begin{table}
	\small
	\begin{center}
		\begin{tabular}{|c|c|c|c|c|c|}
			\hline
			$reg$ & $acc_O$ & $ratio$ & $paras(M)$ & $acc_P$ & $acc_R$ \\
			\hline\hline
			0.0001&91.98&17.48&2.98&90.54&92.71 \\ \hline
			0.0003&90.76&32.56&1.93&89.04&91.94 \\ \hline
			0.0005&89.37&43.93&1.29&54.76&90.93 \\ \hline
			$0.00065^*$&88.35&48.58&1.07&66.41&90.42 \\ \hline
			0.0007&87.98&53.38&0.89&49.52&89.85 \\ \hline
			0.0008&87.19&55.89&0.78&53.95&89.41 \\ \hline
			0.001 &85.18&66.50&0.49&29.91&87.78 \\
			\hline
		\end{tabular}
	\end{center}
	
	\caption{Results of different batch norm scale regularization factors. $reg$ means the value of scale regularization factor, $acc_O$ and $acc_P$ mean the accuracy before and after pruning. $ratio$ is the ratio of pruned channels, $paras$ is the number of parameters in the pruned network. $acc_R$ is the accuracy that we retrain the pruned network from scratch. We use threshold $0.001$ for $0.00065^*$.}
	\label{Tab.3}
	\vspace{-1em}
\end{table}

As the pruned result of batch norm scale regularization factor $0.00065$ has the same number of parameters with $width$$=$$2$ networks, we try to improve its accuracy with knowledge distillation method. The full precision network with accuracy ($92.19\%$) and $width$$=$$8$ binary network ($94.22\%$) are chosen as teacher model, respectively. The values in the full connection layer are used as Hinton et al.~\cite{Distill}. Results are shown in Table~\ref{Tab.4}. With knowledge distillation, the accuracy of pruned binary network improves by $1\%$ and gets $91.39\%$ which is close to the full precision $92.19\%$. The result of $width$$=$$8$ teacher has higher accuracy in contrast to full precision teacher.

\begin{table}
	\begin{center}
		\begin{tabular}{|c|c|c|c|}
			\hline
			$teacher$ & $\tau$ & $\mu$ & $acc$ \\
			\hline\hline
			full&3&0.2&90.72 \\ \hline
			full&5&0.3&91.0 \\ \hline
			full&10&0.2&91.12 \\ \hline
			$wd$$=$$8$&3&0.2&90.91 \\ \hline
			$wd$$=$$8$&5&0.3&91.22 \\ \hline
			$wd$$=$$8$&10&0.2&91.39 \\
			\hline
		\end{tabular}
	\end{center}
	
	\caption{Results of knowledge distillation for CIFAR-10. $wd$$=$$8$ means the teacher is $width$$=$$8$ binary network. $\tau$ is the temperature and $\mu$ is the balance item between knowledge distillation and cross entropy.}
	\label{Tab.4}
\end{table}

In conclusion, by taking use of $2$ times parameters, the performance of binary ResNet-20 is very close to full precision ResNet-20.

\subsection{CIFAR-100}

$n$$=$$1$ bit and $n$$=$$4$ bit quantization ResNet-20 are tested. In Table~\ref{Tab.5}, the baseline ($n$$=$$32$ bit) top-1 accuracy is $69.78\%$. For $n$$=$$1$ bit binary network, the accuracy become higher with the increase of network width. When $width$$=$$4$, the accuracy $70.45\%$ of binary network surpasses the baseline. For $n$$=$$4$ bit quantization network, the accuracy $70.25\%$ surpasses the baseline when $width$$=$$2$. This result shows that more features are needed when quantization networks get less bits.

\begin{table}
	\begin{center}
		\begin{tabular}{|c|c|c|c|}
			\hline
			$bits$ & $width$ & $paras(M)$ & $accuracy$  \\
			\hline\hline
			32   & 1 & 0.28& 69.78 \\
			\hline
			1    & 1 & 0.28& 50.44 \\
			\hline
			1    & 2 & 1.08&62.62 \\
			\hline
			1    & 3 & 2.43&67.61 \\
			\hline
			1    & 4 & 4.31&70.45 \\
			\hline
			1    & 8 & 17.17&74.68 \\
			\hline
			4    & 1 & 0.28&63.35 \\
			\hline
			4    & 2 & 1.08&70.25 \\
			\hline
			4    & 4 & 4.31&73.85 \\
			\hline
		\end{tabular}
	\end{center}
	
	\caption{Results of different quantization bits and widths on CIFAR-100.}
	\label{Tab.5}
	\vspace{-1.0em}
	
\end{table}

\begin{figure}[t]
	\begin{center}
		\includegraphics[width=1.0\linewidth]{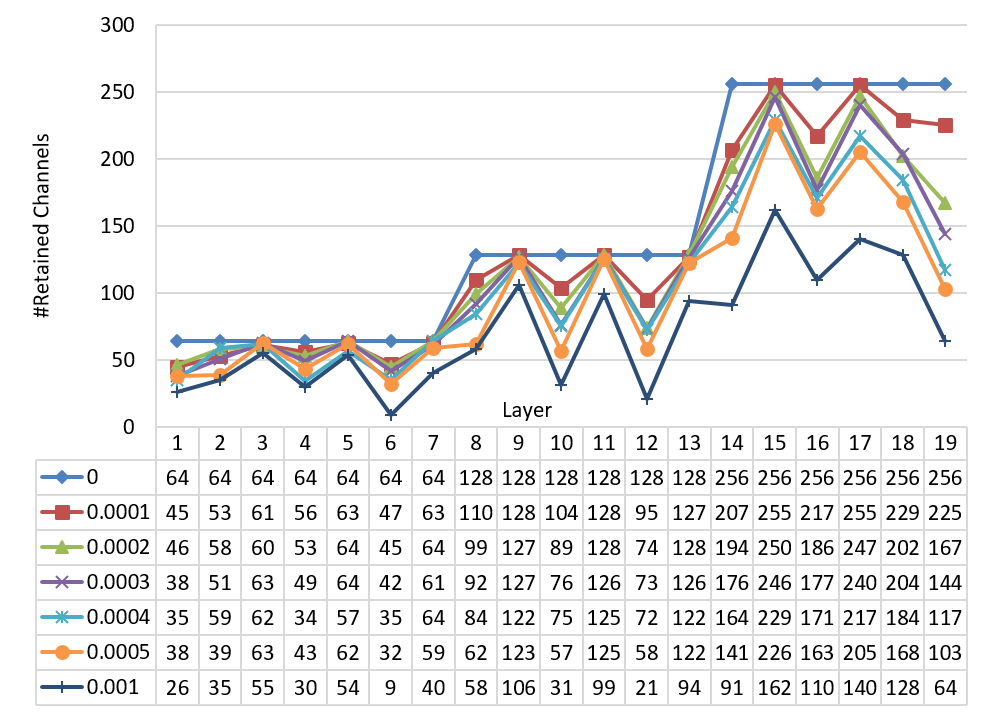}
	\end{center}
	\caption{Network slimming results with different batch norm scale regularization $\lambda$ on CIFAR-100.}
	\label{fig:cifar100_slimming}
\end{figure}

Then we take use of network slimming method to prune the widened networks. Table~\ref{Tab.6} and Figure~\ref{fig:cifar100_slimming} illustrate the pruned network and their accuracy. In contrast to CIFAR-10, CIFAR-100 contains more categories. So more parameters and features are wanted in the pruned network. The threshold of prune is set to $0.01$ in all experiments. When the batch norm scale regularization factor is $0.0005$, the pruned binary ResNet-20 gets $67.98\%$ top-1 accuracy and this network has less parameters in contrast to the $width$$=$$3$ binary ResNet-20. 

\begin{table}
	\begin{center}
		\begin{tabular}{|c|c|c|c|c|c|}
			\hline
			$reg$ & $acc_O$ & $ratio$ & $paras(M)$ & $acc_P$ & $acc_R$ \\
			\hline\hline
			0.0001&69.12&10.32&3.54&66.57&69.54 \\ \hline
			0.0002&67.64&17.11&2.98&59.63&69.36 \\ \hline
			0.0003&67.36&20.97&2.73&61.9&68.46 \\ \hline
			0.0004&65.99&26.31&2.37&61.24&68.28 \\ \hline
			0.0005&64.87&31.36&2.03&57.66&67.98 \\ \hline
			0.001 &57.07&50.84&0.99&24.62&62.07 \\
			\hline
		\end{tabular}
	\end{center}
	
	\caption{Results of different batch norm scale regularization factors on CIFAR-100. $reg$ means the value of scale regularization factor, $acc_O$ and $acc_P$ mean the accuracy before and after pruning. $ratio$ is the ratio of pruned channels, $paras$ is the number of parameters in the pruned network. $acc_R$ is the accuracy that we retrain the pruned network from scratch.}
	\label{Tab.6}
	\vspace{-1.0em}
\end{table}

Knowledge distillation is used to improve the performance of pruned model further. $width=8$ binary ResNet-20 with accuracy $74.68\%$ is chosen as the teacher. Temperature $\tau$ and balance item are set to $3.0$ and $0.8$, respectively. The top-1 accuracy improves to $70.52\%$ which is higher than the baseline full precision accuracy $69.78\%$. Besides, we fine-tune the retrained binary ResNet-20 by frozen all trainable parameters except the batch norm layer. The accuracy improves to $71.38\%$ further.

\subsection{ImageNet}

For ImageNet, we only did experiments on binary network ResNet-18 and VGG16. Firstly, the results of widened networks are shown in Table~\ref{Tab.7} and Table~\ref{Tab.8}. With the increase of width, the batch size become smaller due to limited GPU memory. For ResNet-18, the accuracy of $width$$=$$5$ binary network is higher than the full precision network. For VGG16, the accuracy of $width$$=$$4$ binary network is comparable to the full precision network. These experiments also verify that more quantized feature is benefit to the improvement of quantization networks performance.

\begin{table}
	\begin{center}
		\begin{tabular}{|c|c|c|c|c|}
			\hline
			$bits$ & $width$ & $batch$ & $top-1$ & $top-5$  \\
			\hline\hline
			32&1&1024&70.79&89.5 \\ \hline
			1&1&1024&52.6&76.84 \\ \hline
			1&2&1024&63.73&85.3 \\ \hline
			1&3&1024&68.07&87.92 \\ \hline
			1&4&1024&69.74&89.05 \\ \hline
			1&5&512 &71.08&89.74 \\
			\hline
		\end{tabular}
	\end{center}
	
	\caption{Results of different width ResNet-18 on ImageNet.}
	\label{Tab.7}
\end{table}

\begin{table}
	\begin{center}
		\begin{tabular}{|c|c|c|c|c|}
			\hline
			$bits$ & $width$ & $batch$ & $top-1$ & $top-5$  \\
			\hline\hline
			32&1&1024&71.41&90.47 \\ \hline
			1&1&1024&65.99&86.57 \\ \hline
			1&2&512&69.85&89.33 \\ \hline
			1&4&256&71.01&90.02 \\
			\hline
		\end{tabular}
	\end{center}
	
	\caption{Results of different width VGG16 on ImageNet.}
	\label{Tab.8}
\end{table}

\begin{figure}[t]
	\begin{center}
		\includegraphics[width=1.0\linewidth]{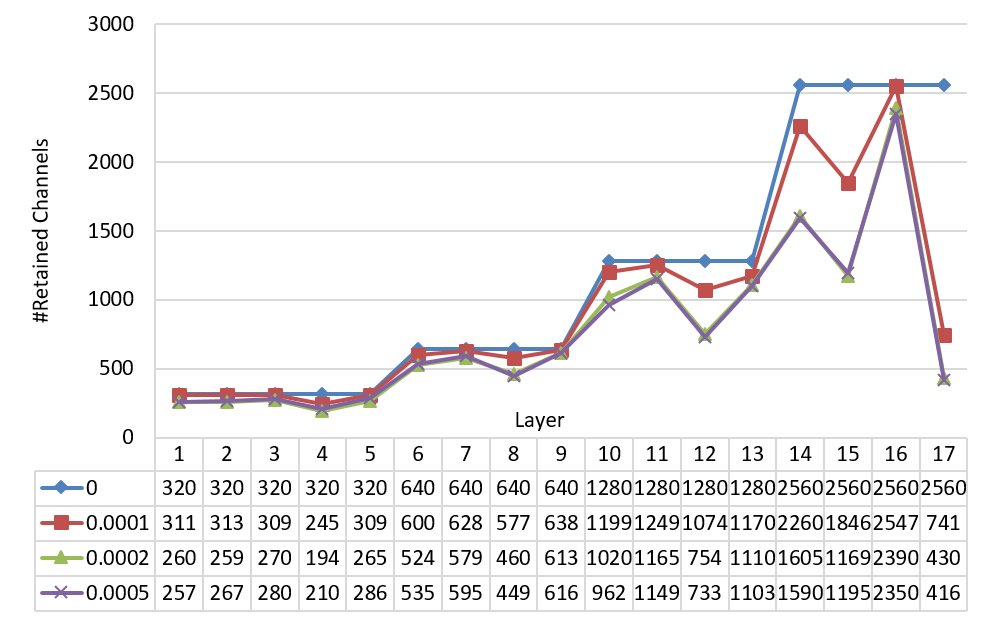}
	\end{center}
	\caption{Network slimming results with different batch norm scale regularization $\lambda$ on ImageNet.}
	\label{fig:imagenet_slimming}
	\vspace{-1em}
\end{figure}

In order to get efficient and accurate quantization network, network slimming and knowledge distillation are used in ResNet-18. In these experiments, the width is set to $5$. Table~\ref{Tab.9} and Figure~\ref{fig:imagenet_slimming} present the results of network slimming. When batch norm scale regularization factor is $0.0005$, the remained channels are about $3.3$ times of the baseline.  Full precision ResNet-18 with accuracy $70.79\%$ is chosen as the teacher. Temperature $\tau$ and balance item are set to $3.0$ and $0.3$, respectively. The top-1 accuracy improves to $70.04\%$ which is comparable to the baseline full precision accuracy.

\begin{table}
	\begin{center}
		\begin{tabular}{|c|c|c|c|c|}
			\hline
			$reg$ & $top1_O$ & $ratio$ & $top1_P$ & $top1_R$ \\
			\hline\hline
			0.0001&68.34&17.95&67.55&68.32 \\ \hline
			0.0002&65.27&33.06&60.84&69.19 \\ \hline
			0.0005&65.10&33.44&58.93&69.18 \\ 
			\hline
		\end{tabular}
	\end{center}
	\caption{Results of different batch norm scale regularization factors on ImageNet. $reg$ means the value of scale regularization factor, $top1_O$ and $top1_P$ mean the accuracy before and after pruning. $ratio$ is the ratio of pruned channels, $top1_R$ is the accuracy that we retrain the pruned network from scratch.}
	\label{Tab.9}
\end{table}

\subsection{Object detection}
To verify the generalization of our method, we further conduct experiments for object detection task. SSD~\cite{SSD} is one-stage object detection method which is widely used due to its efficiency and high accuracy. We use $1$-bit neural network to replace the backbone, extras convolutional layers and detection head. The results are shown in Table~\ref{Tab.10}, where the mean Average Precision (mAP) is adopted as evaluation metric. The baseline mAP is got by SSD with full precision VGG16 backbone. Method $A$ and $C$ replace only the full precision VGG16 backbone with $width$$=$$1$ and $width$$=$$2$ binary VGG16 backbone, respectively. For method $B$ and $D$, we replace all full precision convolutional layers with binary convolutional layers except the first input convolutional layer and the last classification and localization layer. From Table~\ref{Tab.10}, even if only use $width$$=$$1$ binary backbone represented by $A$, we can get $68.74\%$ mAP. In case $B$, the mAP only decreases $1.3$ percent to $67.44\%$. SSD with $width$$=$$2$ binary VGG16 backbone gets $72.79\%$ mAP. These experiments prove that more quantization features are still helpful to improve the performance for object detection task.

\begin{table}
	\begin{center}
		\begin{tabular}{|l|c|}
			\hline
			$method$ & mAP (\%) \\
			\hline\hline
			Baseline FP32 SSD &76.81 \\ \hline
			A: Binary backbone 1$\times$ &68.74 \\ \hline
			B: Binary backbone \& head 1$\times$ &67.44 \\ \hline
			C: Binary backbone 2$\times$ &72.79 \\ \hline
			D: Binary backbone \& head 2$\times$ &71.75 \\
			\hline
		\end{tabular}
	\end{center}
	\caption{Results of different quantization method on PASCAL VOC07 test dataset.}
	\label{Tab.10}
	\vspace{-1.0em}
\end{table}

\section{Conclusion}

To improve the performance of quantization networks, one efficient quantization network design and training method is presented in this paper. The quantization of full precision features will lead to critical information loss and low accuracy. In order to improve this, we take use of network pruning method on widened networks to search for efficient and accurate quantized model. Beside, knowledge distillation is used to improve the performance further. Experiments conducted on benchmark models and datasets verify the effectiveness of proposed method and the experiment results get comparable performance with full precision models. In addition, the proposed method can be combined with latest ideas such as bi-real net, new quantized activation functions and values to get more effective and accurate quantized networks.

{\small
\bibliographystyle{ieee_fullname}
\bibliography{egbib}

\begin{thebibliography}{10}\itemsep=-1pt

\bibitem{deeplab_v1}
Liang-Chieh Chen, George Papandreou, Iasonas Kokkinos, Kevin Murphy, and
  Alan~L. Yuille.
\newblock Semantic image segmentation with deep convolutional nets and fully
  connected crfs.
\newblock {\em arXiv preprint arXiv:1412.7062}, 2014.

\bibitem{binaryconnect}
Matthieu Courbariaux, Yoshua Bengio, and Jean-Pierre David.
\newblock Binaryconnect: Training deep neural networks with binary weights
  during propagations.
\newblock In {\em NIPS}, 2015.

\bibitem{binary}
Matthieu Courbariaux, Itay Hubara, Daniel Soudry, Ran El-Yaniv, and Yoshua
  Bengio.
\newblock Binarized neural networks: Training deep neural networks with weights
  and activations constrained to+ 1 or-1.
\newblock {\em arXiv preprint arXiv:1602.02830}, 2016.

\bibitem{bnn+}
Sajad Darabi, Mouloud Belbahri, Matthieu Courbariaux, and Vahid~Partovi Nia.
\newblock Bnn+: Improved binary network training.
\newblock {\em arXiv preprint arXiv:1812.11800}, 2018.

\bibitem{pascal}
Mark Everingham and John Winn.
\newblock The pascal visual object classes challenge 2012 (voc2012) development
  kit.
\newblock {\em Pattern Analysis, Statistical Modelling and Computational
  Learning, Tech. Rep}, 2011.

\bibitem{RCNN}
Ross Girshick, Jeff Donahue, Trevor Darrell, and Jitendra Malik.
\newblock Rich feature hierarchies for accurate object detection and semantic
  segmentation.
\newblock In {\em CVPR}, 2014.

\bibitem{pruning}
Song Han, Huizi Mao, and William~J Dally.
\newblock Deep compression: Compressing deep neural networks with pruning,
  trained quantization and huffman coding.
\newblock In {\em ICLR}, 2016.

\bibitem{pruning15}
Song Han, Jeff Pool, John Tran, and William Dally.
\newblock Learning both weights and connections for efficient neural network.
\newblock In {\em NIPS}, 2015.

\bibitem{prun1}
Stephen~Jos{\'e} Hanson and Lorien Pratt.
\newblock Comparing biases for minimal network construction with
  back-propagation.
\newblock In {\em NIPS}, 1989.

\bibitem{prun2}
Babak Hassibi and David~G Stork.
\newblock Second order derivatives for network pruning: optimal brain surgeon.
\newblock In {\em NIPs}, 1993.

\bibitem{maskrcnn}
Kaiming He, Georgia Gkioxari, Piotr Dollar, and Ross Girshick.
\newblock Mask r-cnn.
\newblock In {\em ICCV}, 2017.

\bibitem{ResNet}
Kaiming He, Xiangyu Zhang, Shaoqing Ren, and Jian Sun.
\newblock Deep residual learning for image recognition.
\newblock In {\em CVPR}, 2016.

\bibitem{LPP}
Xiaofei He and Partha Niyogi.
\newblock Locality preserving projections.
\newblock In {\em Advances in neural information processing systems}, pages
  153--160, 2004.

\bibitem{softprune}
Yang He, Guoliang Kang, Xuanyi Dong, Yanwei Fu, and Yi Yang.
\newblock Soft filter pruning for accelerating deep convolutional neural
  networks.
\newblock {\em arXiv preprint arXiv:1808.06866}, 2018.

\bibitem{filterprune}
Yang He, Ping Liu, Ziwei Wang, Zhilan Hu, and Yi Yang.
\newblock Filter pruning via geometric median for deep convolutional neural
  networks acceleration.
\newblock In {\em Proceedings of the IEEE Conference on Computer Vision and
  Pattern Recognition}, pages 4340--4349, 2019.

\bibitem{channelprune}
Yihui He, Xiangyu Zhang, and Jian Sun.
\newblock Channel pruning for accelerating very deep neural networks.
\newblock In {\em Proceedings of the IEEE International Conference on Computer
  Vision}, pages 1389--1397, 2017.

\bibitem{Distill}
Geoffrey Hinton, Oriol Vinyals, and Jeff Dean.
\newblock Distilling the knowledge in a neural network.
\newblock {\em arXiv preprint arXiv:1503.02531}, 2015.

\bibitem{mobilenet_v1}
Andrew~G Howard, Menglong Zhu, Bo Chen, Dmitry Kalenichenko, Weijun Wang,
  Tobias Weyand, Marco Andreetto, and Hartwig Adam.
\newblock Mobilenets: Efficient convolutional neural networks for mobile vision
  applications.
\newblock {\em arXiv preprint arXiv:1704.04861}, 2017.

\bibitem{quan2}
Kyuyeon Hwang and Wonyong Sung.
\newblock Fixed-point feedforward deep neural network design using weights+ 1,
  0, and- 1.
\newblock In {\em IEEE Workshop on Signal Processing Systems}, 2014.

\bibitem{squeezenet}
Forrest~N Iandola, Song Han, Matthew~W Moskewicz, Khalid Ashraf, William~J
  Dally, and Kurt Keutzer.
\newblock Squeezenet: Alexnet-level accuracy with 50x fewer parameters and< 0.5
  mb model size.
\newblock {\em arXiv preprint arXiv:1602.07360}, 2016.

\bibitem{matrix1}
Yong-Deok Kim, Eunhyeok Park, Sungjoo Yoo, Taelim Choi, Lu Yang, and Dongjun
  Shin.
\newblock Compression of deep convolutional neural networks for fast and low
  power mobile applications.
\newblock In {\em ICLR}, 2016.

\bibitem{cifar10}
Alex Krizhevsky, Geoffrey Hinton, et~al.
\newblock Learning multiple layers of features from tiny images.
\newblock Technical report, Citeseer, 2009.

\bibitem{AlexNet}
Alex Krizhevsky, Ilya Sutskever, and Geoffrey~E Hinton.
\newblock Imagenet classification with deep convolutional neural networks.
\newblock In {\em NIPS}, 2012.

\bibitem{matrix3}
Vadim Lebedev, Yaroslav Ganin, Maksim Rakhuba, Ivan Oseledets, and Victor
  Lempitsky.
\newblock Speeding-up convolutional neural networks using fine-tuned
  cp-decomposition.
\newblock In {\em ICLR}, 2015.

\bibitem{LeNet}
Yann LeCun, L{\'e}on Bottou, Yoshua Bengio, and Patrick Haffner.
\newblock Gradient-based learning applied to document recognition.
\newblock {\em Proceedings of the IEEE}, 86(11):2278--2324, 1998.

\bibitem{li2016ternary}
Fengfu Li, Bo Zhang, and Bin Liu.
\newblock Ternary weight networks, 2016.

\bibitem{li2016pruning}
Hao Li, Asim Kadav, Igor Durdanovic, Hanan Samet, and Hans~Peter Graf.
\newblock Pruning filters for efficient convnets.
\newblock {\em arXiv preprint arXiv:1608.08710}, 2016.

\bibitem{ABCnet}
Xiaofan Lin, Cong Zhao, and Wei Pan.
\newblock Towards accurate binary convolutional neural network.
\newblock In {\em Advances in Neural Information Processing Systems}, pages
  345--353, 2017.

\bibitem{liu2019learning}
Chuanjian Liu, Yunhe Wang, Kai Han, Chunjing Xu, and Chang Xu.
\newblock Learning instance-wise sparsity for accelerating deep models.
\newblock In {\em Proceedings of the 28th International Joint Conference on
  Artificial Intelligence}, pages 3001--3007. AAAI Press, 2019.

\bibitem{SSD}
Wei Liu, Dragomir Anguelov, Dumitru Erhan, Christian Szegedy, Scott Reed,
  Cheng-Yang Fu, and Alexander~C Berg.
\newblock Ssd: Single shot multibox detector.
\newblock In {\em ECCV}, 2016.

\bibitem{netslimming}
Zhuang Liu, Jianguo Li, B Shen, Gao Huang, Shoumeng Yan, and Changshui Zhang.
\newblock Learning efficient convolutional networks through network slimming.
\newblock In {\em Proceedings of the IEEE International Conference on Computer
  Vision}, pages 2736--2744, 2017.

\bibitem{bireal}
Zechun Liu, Baoyuan Wu, Wenhan Luo, Xin Yang, Wei Liu, and Kwang-Ting Cheng.
\newblock Bi-real net: Enhancing the performance of 1-bit cnns with improved
  representational capability and advanced training algorithm.
\newblock In {\em Proceedings of the European Conference on Computer Vision
  (ECCV)}, pages 722--737, 2018.

\bibitem{FCN}
Jonathan Long, Evan Shelhamer, and Trevor Darrell.
\newblock Fully convolutional networks for semantic segmentation.
\newblock In {\em CVPR}, 2015.

\bibitem{XNORNet}
Mohammad Rastegari, Vicente Ordonez, Joseph Redmon, and Ali Farhadi.
\newblock Xnor-net: Imagenet classification using binary convolutional neural
  networks.
\newblock In {\em ECCV}, 2016.

\bibitem{fasterRCNN}
Shaoqing Ren, Kaiming He, Ross Girshick, and Jian Sun.
\newblock Faster r-cnn: Towards real-time object detection with region proposal
  networks.
\newblock In {\em NIPS}, 2015.

\bibitem{FitNet}
Adriana Romero, Nicolas Ballas, Samira~Ebrahimi Kahou, Antoine Chassang, Carlo
  Gatta, and Yoshua Bengio.
\newblock Fitnets: Hints for thin deep nets.
\newblock {\em arXiv preprint arXiv:1412.6550}, 2014.

\bibitem{LLE}
Sam~T Roweis and Lawrence~K Saul.
\newblock Nonlinear dimensionality reduction by locally linear embedding.
\newblock {\em science}, 290(5500):2323--2326, 2000.

\bibitem{ImageNet}
Olga Russakovsky, Jia Deng, Hao Su, Jonathan Krause, Sanjeev Satheesh, Sean Ma,
  Zhiheng Huang, Andrej Karpathy, Aditya Khosla, Michael Bernstein, et~al.
\newblock Imagenet large scale visual recognition challenge.
\newblock {\em IJCV}, 115(3):211--252, 2015.

\bibitem{shen2019searching}
Mingzhu Shen, Kai Han, Chunjing Xu, and Yunhe Wang.
\newblock Searching for accurate binary neural architectures.
\newblock In {\em Proceedings of the IEEE International Conference on Computer
  Vision Workshops}, pages 0--0, 2019.

\bibitem{VGGnet}
Karen Simonyan and Andrew Zisserman.
\newblock Very deep convolutional networks for large-scale image recognition.
\newblock {\em ICLR}, 2015.

\bibitem{GoogleNet}
Christian Szegedy, Wei Liu, Yangqing Jia, Pierre Sermanet, Scott Reed, Dragomir
  Anguelov, Dumitru Erhan, Vincent Vanhoucke, and Andrew Rabinovich.
\newblock Going deeper with convolutions.
\newblock In {\em CVPR}, 2015.

\bibitem{isomap}
Joshua~B Tenenbaum, Vin De~Silva, and John~C Langford.
\newblock A global geometric framework for nonlinear dimensionality reduction.
\newblock {\em science}, 290(5500):2319--2323, 2000.

\bibitem{quan1}
Vincent Vanhoucke, Andrew Senior, and Mark~Z Mao.
\newblock Improving the speed of neural networks on cpus.
\newblock In {\em Deep Learning and Unsupervised Feature Learning Workshop,
  NIPS}, 2011.

\bibitem{CBE}
Felix Yu, Sanjiv Kumar, Yunchao Gong, and Shih-Fu Chang.
\newblock Circulant binary embedding.
\newblock In {\em International conference on machine learning}, pages
  946--954, 2014.

\bibitem{attentiontransfer}
Sergey Zagoruyko and Nikos Komodakis.
\newblock Paying more attention to attention: Improving the performance of
  convolutional neural networks via attention transfer.
\newblock {\em arXiv preprint arXiv:1612.03928}, 2016.

\bibitem{LQNet}
Dongqing Zhang, Jiaolong Yang, Dongqiangzi Ye, and Gang Hua.
\newblock Lq-nets: Learned quantization for highly accurate and compact deep
  neural networks.
\newblock In {\em Proceedings of the European Conference on Computer Vision
  (ECCV)}, pages 365--382, 2018.

\bibitem{shufflenet_v1}
Xiangyu Zhang, Xinyu Zhou, Mengxiao Lin, and Jian Sun.
\newblock Shufflenet: An extremely efficient convolutional neural network for
  mobile devices.
\newblock In {\em CVPR}, 2018.

\bibitem{dorefanet}
Shuchang Zhou, Yuxin Wu, Zekun Ni, Xinyu Zhou, He Wen, and Yuheng Zou.
\newblock Dorefa-net: Training low bitwidth convolutional neural networks with
  low bitwidth gradients, 2016.

\bibitem{TTQ}
Chenzhuo Zhu, Song Han, Huizi Mao, and William~J. Dally.
\newblock Trained ternary quantization, 2016.

\bibitem{binaryensemble}
Shilin Zhu, Xin Dong, and Hao Su.
\newblock Binary ensemble neural network: More bits per network or more
  networks per bit?
\newblock In {\em Proceedings of the IEEE Conference on Computer Vision and
  Pattern Recognition}, pages 4923--4932, 2019.

\end{thebibliography}
}

\end{document}